\documentclass[10pt]{article}
\usepackage{times}
\usepackage[none]{hyphenat}
\usepackage{url}
\usepackage[utf8]{inputenc}
\usepackage{graphicx}

\usepackage{multirow}
\usepackage{booktabs}
\usepackage{tcolorbox}
\usepackage{hyperref}
\usepackage{authblk}

\title{Energy Efficiency of Training Neural Network Architectures: An Empirical Study}

\author[1]{Yinlena Xu}
\author[1]{Silverio Martínez-Fernández}
\author[2]{Matias Martinez}
\author[1]{ Xavier Franch}

\affil[1]{Universitat Politècnica de Catalunya}
\affil[2]{Université Polytechnique Hauts-de-France}

\date{}

\begin{document}
\maketitle
\begin{abstract}
The evaluation of Deep Learning (DL) models has traditionally focused on criteria such as accuracy, F1 score, and related measures. The increasing availability of high computational power environments allows the creation of deeper and more complex models. However, the computations needed to train such models entail a large carbon footprint. In this work, we study the relations between DL model architectures and their environmental impact in terms of energy consumed and CO$_2$ emissions produced during training by means of an empirical study using Deep Convolutional Neural Networks. Concretely, we study: (\textit{i}) the impact of the architecture and the location where the computations are hosted on the energy consumption and emissions produced; (\textit{ii}) the trade-off between accuracy and energy efficiency; and (\textit{iii}) the difference on the method of measurement of the energy consumed using software-based and hardware-based tools.
\end{abstract}

\subsubsection*{Keywords:}

Green AI, deep learning, neural networks, sustainable software engineering, energy metrics

\section{Introduction}\label{sec:intriduction}

In recent years, Deep Learning (DL) models have shown great performance in many machine learning-based tasks. The DL-centric research paradigm and the ambition of creating the next state-of-the-art model lead to the exponential growth of model size and the use of larger datasets to train these models, requiring therefore intensive computation that entails a considerable large financial cost and carbon footprint \cite{strubell2019energy}. If this trend continues, greater amounts of energy will be needed to build larger models to achieve ever-smaller improvements, making research progress directly depend on the uncontrolled exploitation of computing resources. In this context, energy consumption is becoming a necessary consideration when designing all types of software \cite{hicss20schmermbeckgreen} and specifically DL-based solutions. Fortunately, the awareness of aligning DL research with the emergent Green AI movement \cite{schwartz2020green} is growing steadily.

In the DL realm, Convolutional Neural Networks (CNN) have become a well-known architectural approach widely used in areas such as image classification and natural language processing (NLP) \cite{lecun1998gradient} \cite{russakovsky2015imagenet}. Common architectures for CNNs are AlexNet \cite{russakovsky2015imagenet}, VGGNet \cite{simonyan2014very}, GoogleNet \cite{szegedy2015going}, and ResNet \cite{he2016deep}. CNNs use linear algebra principles, specifically matrix multiplication, to identify patterns. Alternative activation function, parameter optimization, and architectural innovations were the basis of CNN advances. These networks are computationally demanding, requiring graphical processing units (GPU) to train the models. The availability of large amounts of data and the access to more powerful hardware has opened new possibilities for CNN research. Indeed, the evolution of these architectures has shown a trend towards increasingly complex models to solve increasingly complex tasks \cite{krizhevsky2012imagenet} \cite{szegedy2015going} \cite{chollet2017xception} \cite{silver2017mastering}.

In this paper, we investigate the effects of different CNN architectures in the energy efficiency of the model training stage, and the possible relation of energy efficiency with the accuracy of the obtained model. To do so, we focus on one particular application domain, namely computer vision (CV), which has evolved significantly in the last years thanks to the widespread application of CNNs. CV applications are useful in many areas including medical imaging, agriculture monitoring, traffic control systems, sports tracking, and more. This includes a set of challenges such as image classification, object detection, image segmentation, image captioning among others. We center this work in the context of image classification as it is considered the basis for CV problems and CNNs have become the state-of-the-art technique. To perform this task, CNNs extract important features from the images at each convolution level and are completed with some fully connected output nodes for the classification.


This document is structured as follows. Section \ref{sec:bg-related-work} gives the background and reports the related work on energy consumption of DL systems and Green AI. Section \ref{sec:research-methodology} defines the research goal and research questions of our study, as well as the experimental methodology. Section \ref{sec:results} presents the results and gives answers to our research questions. Section \ref{sec:discussion} reviews findings and discuss their implications and Section \ref{sec:conclusions} summarizes the overall study and delineates future steps.

\section{Background and Related Work}\label{sec:bg-related-work}


\subsection{Energy measurement}

To measure the energy consumption of a computing device there are essentially two kinds of tools: hardware power monitors and energy profilers \cite{cruz2021tools}. Hardware monitors are directly connected to the power source of the component that can be used to monitor the energy consumption of software. Despite being difficult to set up, power monitors are the most accurate strategy to measure energy, although they cannot discriminate what percentage of this consumption comes from a particular thread of execution. The other strategy is using energy profilers, a software-based tool that captures energy data in conjunction to program execution. This allows energy profilers to compute the power consumed by the device, but these calculation rely on estimations. To what extent these estimations differ from the real consumption is worth to be investigated in order to claim for internal validity of empirical studies on energy efficiency.

Recent work has analyzed the carbon footprint of training deep learning models and advocated for the evaluation of the energy efficiency as an evaluation criterion for research \cite{schwartz2020green}. The number of floating point operations (FLOPs) has been used in the past to quantify the energy footprint of a model \cite{vaswani2017attention} \cite{molchanov2016pruning} \cite{liu2021swin}, but they are not widely adopted in DL research. And little research has been done regarding the CO$_2$ emissions of highly expensive computation processes.

\subsection{Green AI}

The present-days concern on the carbon footprint of increasingly large DL models has been growing. Schwartz et al. \cite{schwartz2020green} advocate for redirecting DL research towards a more environmentally friendly solution known as Green AI. They estimated that computational cost of AI research that aim to obtain state-of-the-art results has increased 300.000x from 2012 to 2018. This is due to the AI community focus on metrics such as accuracy rather than energy efficiency. In this paper, they suggest to report the number of FLOPs required to generate the results as a standard measure of efficiency.

Strubell et al. \cite{strubell2019energy} estimated the carbon emission of training some of the recently successful neural network models for NLP, raising awareness and proposing actionable recommendations to reduce costs of NLP research. They conclude that these trends are not only found in the NLP community, but hold true across the AI community in general.

Recent work by Google and UC Berkeley \cite{patterson2021carbon} has estimated the carbon footprint and energy consumption of large neural network training. The paper proposes strategies to improve the energy efficiency and CO$_2$ emissions. They reported that by carefully choosing processor, hardware and data centers, it is possible to reduce the carbon footprint of deep neural networks by up to ~100-1000 times.

When it comes to DL frameworks, Georgiou et al. \cite{georgiou2022green} reported clear difference between energy consumption and run-time performance of two of the most popular DL frameworks, Pytorch and Tensorflow. The study showed that DL frameworks show significant model-sensitivity and that current documentation of the frameworks has to be improved. Also, Creus et al. studied how to make greener DL-based mobile applications. The studies showed that it is possible to build optimized DL-based applications varying the number of paramenters of CNNs \cite{castanyer2021design,castanyer2021integration}.

Regarding greener DL models in applications domains, we can find advances for greener DL-based solutions as well. For instance, for weed detection, Ofori et al. combined the mobile-sized EfficientNet with transfer learning to achieve up to 95.44\% classification accuracy on plant seedlings \cite{hicss21ofori2021approach}, and model compression achieving 62.22\% smaller in size than DenseNet (the smallest-sized full-sized model) \cite{hicss22ofori2022deep}. Moreover, for pig posture classification, Witte et al. reported the YOLOv5 model achieving an accuracy of 99,4\% for pig detection, and EfficientNet achieving a precision of 93\% for pig posture classification \cite{hicss22witte2022introducing}. 

With respect to the aforementioned works, there is a clear need for further research to build greener DL-based solutions and models. Our work delves into CNN architectures and their energy efficiency.



\section{Research methodology}\label{sec:research-methodology}

\subsection{Goal, research questions and hypotheses}\label{sec:research-goal-and-questions}

We formulate our research goal according to the Goal Question Metric (GQM) guidelines \cite{caldiera1994goal} as follows:
Analyze \textit{convolutional neural networks architectures} with the purpose of \textit{measuring their energy efficiency} with respect to \textit{the model training} from the point of view of \textit{the AI practitioner} in the context of \textit{creating an image classification model for computer vision}.
This research goal is operationalised into three research questions (RQ):

\textbf{RQ1}: Does the CNN architecture have an impact on energy consumption? According to the background introduced in Section \ref{sec:bg-related-work}, we will respond this RQ using two different measures which generate a null hypothesis each:  
H.1.1.0: There is no difference in energy consumed and emissions produced during training varying the CNN model architecture.
H.1.2.0: There is no difference in FLOPs required during training varying the CNN model architecture.

\textbf{RQ2}: What is the relationship between CNN accuracy and the energy needed to train the model?

\textbf{RQ3}: What are the differences between software-based and hardware-based methods of measuring the energy efficiency of a model?

With RQ1 we aim to provide a comparative analysis of the measures specified for some of the best known CNN architectures for image classification, namely VGG16, VGG19 and ResNet50, to determine the correlation between architecture complexity and energy consumption. We carry out this analysis on two of the most popular image datasets for this task: MNIST and CIFAR-10. 

With RQ2 we want to compare the trade-off between energy efficiency and the accuracy obtained from each model configuration. If little accuracy gains require much more computation, one can argue that this improvement is only needed when facing critical business cases (e.g., designing life-critical systems). To answer this RQ, we will be using a score ratio introduced by Alyamkin et al.  \cite{alyamkin2019low}, where they introduce this new metric to compare models: $Score = Accuracy / Energy$. Energy in our case will refer to the energy consumption of the training in kWh.

With RQ3 we pretend to study how to measure the energy efficiency of a model's training while exploring two different ways of measurement (see Section \ref{sec:bg-related-work}): the use of wattmeters (hardware-based measurement) and the use of profilers (software-based estimation). Understanding the internals of these measurement instruments will help researchers to design robust study protocols.

\subsection{Study Design}

\begin{figure*}
\begin{center}
    \includegraphics[width=\textwidth]{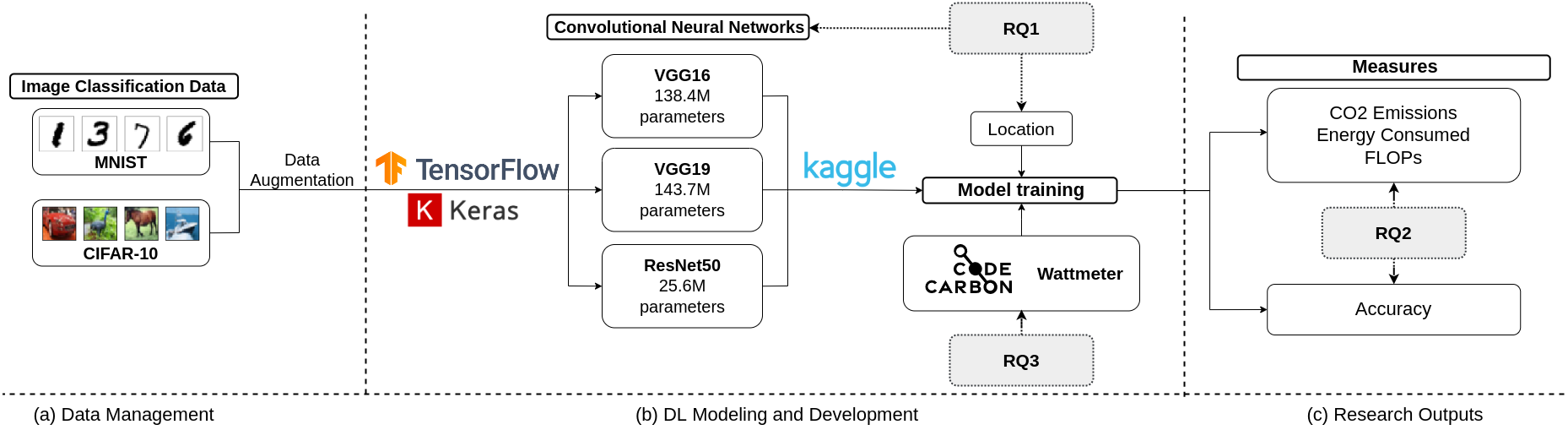}
\end{center}
\caption{Schema of the empirical study.}
\label{fig:overview}
\end{figure*}

We divide the study into a three-stage pipeline (see Fig. \ref{fig:overview}):
(a) the \textit{Data Management} stage which includes the collection and preprocessing of the images, 
(b) the \textit{Modeling and Development} of the DL components, including the training of the DL model, and 
(c) the \textit{Research Outputs}, which studies the outputs from the previous phase (e.g., power, energy consumed, accuracy from the models) to answer our RQs.

\subsection{Variables}\label{sec:variables}

In the following subsections we define the variables of our experimental design grouped into three categories.

\subsubsection{Independent variables.}\label{sec:independent-variables}

In this study we define two independent variables: (\textit{i}) the CNN architecture, and (\textit{ii}) the measurement instrument.

As defined in Section \ref{sec:research-goal-and-questions}, our objective is focused on the energy consumption of training a deep CNN model, and not on the model itself. Therefore we use transfer learning from the following CNN architectures: VGG16, VGG19, and ResNet50. We define the model architecture as a categorical variable that specifies which of the CNN is trained, and the model number of parameters is defined as a numerical variable indicating the complexity of the model.

VGG comes from the Visual Geometry Group from Oxford and it was used to win the ILSVR (ImageNet) competition in 2014 \cite{simonyan2014very}.

VGG16 is a 16-layer model, being 13 of them convolutional and the other 3, fully connected. It has 138.4 million parameters. The same size is used for all the kernels in every convolutional layer is used, namely 3x3 kernel with stride = 1 and padding = 1. For maximum pooling, this changes into 2x2 kernel with stride = 2. 

VGG19 is a newer version, build upon the same concept as the VGG16, but with 19 layers in total, with 16 convolutional layers and 143.7M million parameters.

ResNet stands for Residual Network and was first introduced in 2015 \cite{he2016deep}. The architecture of this NN relies on Residual Blocks, where a residual block is a combination of the original input and an output after convolution and activation function. There are different versions of ResNet having a different number of layers. 

ResNet50 is the 50-layer model that has 48 convolution layers and 25.6 million parameters.

The way of measuring the energy consumption that indicates the two options of getting the measurements: an emission profiler
and a  wattmeter (see Section  \ref{sec:data_collection} for more details).

\subsubsection{Dependent variables.}\label{sec:dependent-variables}

To measure the environmental impact of a model's training process, we will track the computer power and energy consumption during the experiments. We use four numerical variables that measure \emph{(i)} the emissions in CO$_2$ equivalents (CO$_2$-eq) in kg;  \emph{(ii)} the energy consumed by the infrastructure in kWh; \emph{(iii)} the number of floating-point operations (FLOP) needed to train the model; and \emph{(iv)} the validation accuracy of the model obtained. 

\subsubsection{Other variables.}\label{sec:other-variables}

We use a categorical variable that indicates which dataset is utilized for the model training. The image classification input datasets used in this paper are the following:

The {MNIST\footnote{\url{http://yann.lecun.com/exdb/mnist/}}} (Modified National Institute of Standards and Technology) handwritten digits \cite{deng2012mnist}. It is a large dataset commonly used in ML for training systems. It consists of 70,000 28 x 28 black and white images with 10 classes: digits from 0 to 9. The images have been normalized and centered in a fixed size and grayscale levels where introduced with anti-aliasing. There are 60,000 images for training and 10,000 for testing.

The {CIFAR-10\footnote{\url{https://www.cs.toronto.edu/~kriz/cifar.html}}} (Canadian Institute For Advanced Research) dataset. It is a set of small labeled images for classification dataset which consists of 60,000 32 x 32 colour images in 10 mutually exclusive classes, with 6,000 images per class. There are 50,000 training images and 10,000 test images and the classes are: airplane, automobile, bird, cat, deer, dog, frog, horse, ship, and truck. These images are challenging to classify due to the varying lighting condition and angles.

Other categorical variables that should be considered independent but we do not have full control over them due to the cloud provider plan are:

The type of hardware. For RQ1 and RQ2 the computations were performed on 2 x Intel® Xeon® Processor 2.00 GHz CPUs and 1 x Tesla P100-PCIE-16GB GPU. For RQ3 8the computations were performed on 80 x Intel® Xeon® E5-2698 v4 @ 2.20GHz CPUs and 8 x Tesla V100-SXM2-32GB GPUs.

The location where the cloud is hosted. Experiments were conducted using {Kaggle\footnote{\url{https://www.kaggle.com/}}} kernels. For RQ1 and RQ2 there were three available locations: Taipei (Taiwan), Oregon (USA), and South Carolina (USA). For RQ3 the infrastructure was located in Île-de-France (France).

\begin{table*}[]
\caption{Independent, dependent and other variables of the study.}
\resizebox{\textwidth}{!}{%
\centering
\begin{tabular}{@{}lllll@{}}
\toprule
\textbf{Class} & \textbf{Name} & \textbf{Description} & \textbf{Scale} & \textbf{Operationalization} \\ \midrule
Independent & Architecture Type & The deep CNN architecture & nominal & See section \ref{sec:independent-variables} \\
 & Measuring instrument & \begin{tabular}[c]{@{}l@{}}Energy measuring method \\ (by hardware or software)\end{tabular} & nominal & See section \ref{sec:independent-variables} \\ \midrule
Dependent & Emissions & \begin{tabular}[c]{@{}l@{}}Carbon dioxide (CO$_2$) emissions, \\ expressed as kilograms of \\ CO$_2$-equivalents (CO$_2$-eq)\end{tabular} & numerical & Profiled \\
 & Energy consumption & \begin{tabular}[c]{@{}l@{}}Net power supply consumed \\ during the compute time, \\ measured as kWh\end{tabular} & numerical & See measuring method \\
 & Floating-point operations & \begin{tabular}[c]{@{}l@{}}Number of floating point \\ operations per second (FLOP)\end{tabular} & numerical & Retrieved from modeling \\
 & Accuracy & \begin{tabular}[c]{@{}l@{}}Validation accuracy \\ obtained after training\end{tabular} & numerical & Retrieved from modeling \\ \midrule
Others  & Dataset & \begin{tabular}[c]{@{}l@{}}The input dataset used to \\ train the models\end{tabular} & nominal & See section \ref{sec:other-variables} \\
 & Hardware & GPU and CPU type & nominal & Profiled \\ 
 & Location & \begin{tabular}[c]{@{}l@{}}Province/State/City where the\\ compute infrastructure is hosted\end{tabular} & nominal & Profiled \\
 \bottomrule
\end{tabular}}
\label{tab:variables}
\end{table*}

\subsection{Data collection}\label{sec:data_collection}

In this section, we respectively report the measures of our study (see Figure \ref{fig:overview}, (c)), and the instruments used to collect them.

\subsubsection{Measures.}

To describe the amount of work that is required to train a model we compute the following measures:

\textbf{CO$_2$ emission} is the quantity that we want to minimize directly. These emissions can be calculated as the product between: (i) carbon intensity of the electricity consumed for computation, quantified as kg of CO$_2$ emitted per kWh of electricity, and (ii) the net power supply consumed by the computational infrastructure, quantified in kWh. Carbon intensity of electricity used is determined by a weighted average of emissions from various energy sources used to generate power, including fossil fuels and renewables. The combination of energy sources is based on the specific location where the computation is hosted.

\textbf{Energy consumed} is related to CO$_2$ emissions, while being independent of time and location. The power supply to the hardware is tracked at frequent time intervals, thus it is highly dependent on the type of hardware utilized.

We executed our experiment three times and we report the median value of energy consumed reported by both the wattmeter and the profiler.

\textbf{FLOPs} is the total number of floating-point operations required to execute a computational process. It estimates the amount of work needed for the process as a deterministic measure, computed by defining the cost of two base operations: addition and multiplication. FLOPs can be estimated given a model instance even before starting the training.

To compute the FLOPs required for the training of the model we use the {keras-flops\footnote{\url{https://github.com/tokusumi/keras-flops}}} package for TensorFlow. All code has been developed in Python (version 3.7.12) and the {Keras} API of TensorFlow\footnote{\url{https://keras.io/}} and all models were trained with a batch size of 32.

\subsubsection{Instruments.}

To conduct the collection of data and the aforementioned variables, we use two different instruments.

First, we used the {CodeCarbon\footnote{\url{https://github.com/mlco2/codecarbon}}} profiler: a Python package that enables us to track emissions in order to estimate the carbon footprint of an experiment.  Internally, CodeCarbon uses RAPL for measuring the energy consumed by the CPU and RAM, and NVIDIA Management Library (NVML) for the energy consumption of the GPU. CodeCarbon also presents the the total energy consumed, which corresponds to the sum of the energy consumption from the CPU, GPU and RAM. The package logs the data of each experiment into an \textit{emissions.csv} file. The logged fields we are interested in are: duration of the compute (in seconds), emissions as CO$_2$-equivalents (in kg), and energy consumed (in kWh). 

Second, for responding RQ3, we replicated the experiment on a machine connected to a wattmeter, therefore being able to compare the energy consumption obtained using both a wattmeter and a profiler. We used a wattmeter from the OmegaWatt vendor, which is able to collect up to 50 measurements per second of power directly from the power supply units.

\subsection{Data analysis}\label{subsec:data-analysis}

In RQ1, we divided the analysis into two different parts, considering two variables: model architecture and input data. In each part we assessed the energy consumption on all the dependent variables (CO$_2$-equivalent emissions, energy consumed, and FLOPs). Within each part, we followed an identical procedure: (1) use violin and box plots to illustrate the distributions for each response variable, comparing between datasets and CNN architectures; (2) assess the correlation coefficient between independent and dependent variables; (3) assess the statistical significance (i.e., $p$-value) of the findings.

We used a point-biserial correlation coefficient to assess the correlation between dependent variables and the input data. Point-biserial correlation is a correlation coefficient used when we have a dichotomous and a continuous variable. It ranges from $-1$ to $+1$, where $-1$ indicates a perfect negative association, $+1$ indicates a perfect positive association and 0 indicates no association. 

To assess the dependent variables with respect to the type of architecture we used Kruskal-Wallis test. Kruskal-Wallis test by rank is a non-parametric alternative to one-way ANOVA test, which extends the two-samples Wilcoxon test in the situation where there are more than two groups. A significant Kruskal–Wallis test indicates that at least one sample stochastically dominates one other sample.

In RQ2, we assess the trade-off between accuracy and energy consumption with the $Score = Accuracy / Energy$ (see Section  \ref{sec:research-goal-and-questions} for more details).
We can easily compare the scores between between experiments by sorting them.

For RQ3, we compared the energy consumption as collected in two different ways: a wattmeter and CodeCarbon. The relationship between the two methods is assessed by computing the Spearman's rank correlation coefficient.


\section{Results}\label{sec:results}

In this section we discuss the quantitative results in response to the RQs and hypotheses presented in \ref{sec:research-goal-and-questions}. The entire analysis was conducted using R language. 

Table \ref{tab:flops-results} contains the summary of the different experiment configurations and its characteristics. 

\begin{table*}[t]
\caption{Experiment characteristics. Dataset size includes both train and test sets; the split proportion can be found in section \ref{sec:independent-variables}. Depth refers to the topological depth of the network, which includes activation layers, batch normalization layers, etc.}
\resizebox{\textwidth}{!}{%
\centering
\begin{tabular}{@{}lllllllll@{}}
\toprule
\textbf{Architecture} & \textbf{Data} & \textbf{\begin{tabular}[c]{@{}l@{}}Image \\ Size\end{tabular}} & \textbf{\begin{tabular}[c]{@{}l@{}}Dataset \\ Size\end{tabular}} & \textbf{Depth} & \textbf{\begin{tabular}[c]{@{}l@{}}Total \\ Parameters\end{tabular}} & \textbf{\begin{tabular}[c]{@{}l@{}}Trainable \\ Parameters\end{tabular}} & \textbf{\begin{tabular}[c]{@{}l@{}}Total \\ FLOPs\end{tabular}} & \textbf{\begin{tabular}[c]{@{}l@{}}Trainable Layer \\ FLOPs\end{tabular}} \\ \midrule
VGG16 & CIFAR10 & 32x32 & 60k & 16 & 33,6M & 18,9M & 21.3 G & 1.21 G \\
 & MNIST & 48x48 & 70k & 16 & 33,6M & 18,9M & 46.3 G & 1.21 G \\ \midrule
VGG19 & CIFAR10 & 32x32 & 60k & 19 & 38,9M & 18,9M & 26.7 G & 1.21 G \\
 & MNIST & 48x48 & 70k & 19 & 38,9M & 18,9M & 58.6 G & 1.21 G \\ \midrule
ResNet50 & CIFAR10 & 32x32 & 60k & 50 & 48,8M & 25,2M & 6.68 G & 1.61 G \\
 & MNIST & 48x48 & 70k & 50 & 73,9M & 50,3M & 16.3 G & 3.22 G \\ \bottomrule
\end{tabular}}
\label{tab:flops-results}
\end{table*}

\subsection{Does CNN architecture have an impact on energy consumption? (RQ1)}\label{subsec:rq2}

\begin{figure}[t]
\begin{center}
    \includegraphics[width=1\linewidth]{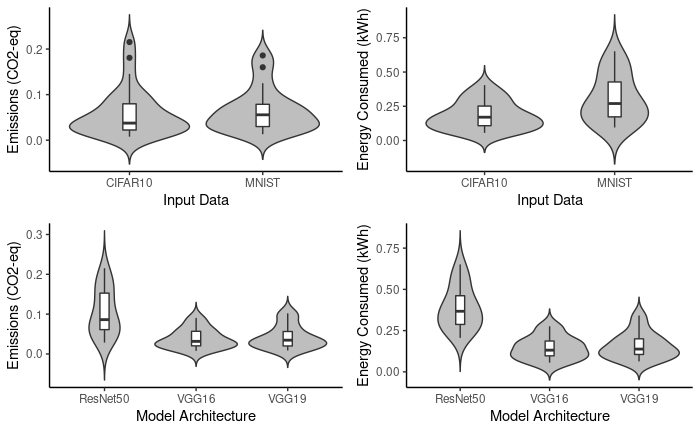}
\end{center}
\caption{Violin-plots for the total emissions and energy consumed with the input dataset and CNN architecture.}
\label{fig:violin-emissions}
\end{figure}

\begin{figure}[t]
\begin{center}
    \includegraphics[width=1\linewidth]{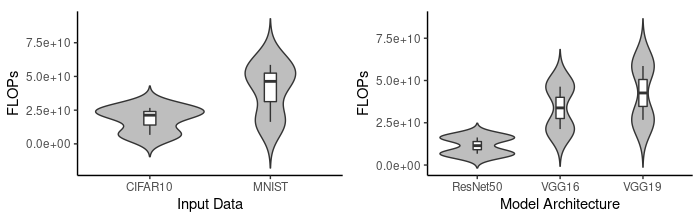}
\end{center}
\caption{Violin-plots for the total FLOPs with the input dataset and CNN architecture.}
\label{fig:violin-flops}
\end{figure}

\begin{figure}[t]
\begin{center}
    \includegraphics[width=1\linewidth]{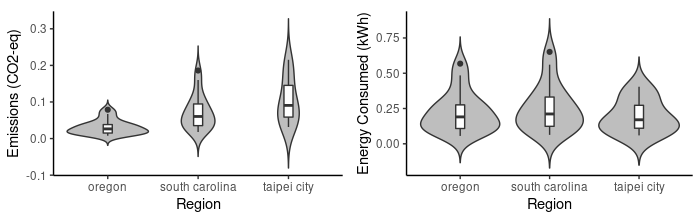}
\end{center}
\caption{Violin-plots for the total emissions and energy consumed with the location where the compute infrastructure is located.}
\label{fig:violin-regions}
\end{figure}

Fig. \ref{fig:violin-emissions} shows the violin plot of the correlation between the energy consumed (in kWh) and the emissions produced (in CO$_2$-eq in kg) with the different experiments grouped by input dataset (CIFAR10 and MNIST) and by CNN architecture (VGG16, VGG19 and ResNet50). The boxplots show that the median of both emissions and energy consumed using the CIFAR10 images is lower than using the MNIST images. We see that regarding the type of architecture, both VGG models report similar emissions and consumed energy, that are lower with respect to ResNet50. In Fig. \ref{fig:violin-flops} have the violin plots for the number of FLOPs required to train the model. In this case we see that using the MNIST dataset and the VGG19 model requires more FLOPs. Furthermore, the same procedure described in section \ref{subsec:data-analysis} for the location variable. Fig. \ref{fig:violin-regions} shows the violin plot and box plots grouped by location. Emissions from Taiwan (Taipei City) were higher than the other two cities located in the United States (Oregon and South Carolina). Regarding the energy consumed, the three locations do not show difference.

\begin{table}[h]
\caption{Assessment for point-biserial correlation coefficient between input datasets and the emission rate, consumed energy and FLOPs.}
\resizebox{\columnwidth}{!}{%
\centering
\begin{tabular}{@{}lll@{}}
\toprule
\textbf{Variable} & \textbf{Corr. Coef.} & \textbf{Assessment} \\ \midrule
Emissions (CO$_2$-eq) & -0.0605 & Weak corr. \\
Energy Consumed (kWh) & -0.4135 & Strong corr. \\
FLOPs & -0.6233 & Strong corr. \\ \bottomrule
\end{tabular}}
\label{tab:point-biserial}
\end{table}

Table \ref{tab:point-biserial} shows the results of computing point-biserial correlation coefficient between the input dataset and the response variables. In this case we take CIFAR10 as the base dataset, meaning that a negative value of the correlation coefficient indicates that the variables are inversely related. We see that both the consumed energy during training and the number of FLOPs are strongly correlated to the input data. Both coefficients are negative, meaning that to train with the CIFAR10 dataset required less energy and FLOPs with respect to using the MNIST dataset.

\begin{table}[h]
\caption{Statistical significance assessment for Kruskal-Wallis test for correlation between the architectures and the emission rate, consumed energy and FLOPs. This tests responds to the hypotheses H.1.1.0 and H.1.2.0 from section \ref{sec:research-goal-and-questions}.}
\resizebox{\columnwidth}{!}{%
\centering
\begin{tabular}{@{}lll@{}}
\toprule
\textbf{Variable} & \textbf{$p$-value} & \textbf{Assessment} \\ \midrule
Emissions (CO$_2$-eq) & $<0.001$ & Significant \\
Energy Consumed (kWh) & $<0.001$ & Significant \\
FLOPs & $0.1561$ & Not significant \\ \bottomrule
\end{tabular}}
\label{tab:kruskal-wallis}
\end{table}

Table \ref{tab:kruskal-wallis} shows the statistic significance (Kruskal-Wallis test $p$-value) between the architecture of the model and the response variables. In summary, there is statistical significance to accept that there is correlation between the emissions produced in CO$_2$-eq and energy consumption with the type of architecture.

\begin{table}[h]
\caption{Statistical significance assessment for Kruskal-Wallis test for correlation between the  infrastructure locations and the emission rate and consumed energy.}
\resizebox{\columnwidth}{!}{%
\centering
\begin{tabular}{@{}lll@{}}
\toprule
\textbf{Variable} & \textbf{$p$-value} & \textbf{Assessment} \\ \midrule
Emissions (CO$_2$-eq) & $<0.001$ & Significant \\
Energy Consumed (kWh) & $0.6509$ & Not significant \\ \bottomrule
\end{tabular}}
\label{tab:kruskal-wallis-regions}
\end{table}

Finally, table \ref{tab:kruskal-wallis-regions} shows the statistical significance (Kruskal-Wallis test $p$-value) between the location where the computation was hosted and the emissions and energy consumed. The $p$-values of the test statistic show that there is relation between the location and the emissions produced to train the model, but not with the energy consumed.





\subsection{
What is the relationship between model accuracy and the energy needed to train the model? (RQ2)
}\label{subsec:rq3}

Table \ref{tab:scores} presents the accuracy obtained from a model with a particular architecture trained on a dataset in a particular location, the energy consumed for training that model, and the Score which correspond to the ratio between the mentioned accuracy and energy.

We first analyze the score by location (as allow us to compare energy consumed by the same hardware on different models).

In Oregon and in South Carolina, we observe that a model with architecture VGG19 trained on CIFAR10 produces the highest Score (12.64). VGG19 has a lower energy consumption at the expense of having a lower accuracy compared to its smoller version VGG16. The score metric allows to quantify the trade-off between energy and accuracy. 

In Taipei, the model with VGG16 architecture has a higher accuracy than VGG19 wih higher energy consumtion (as happened in the previous two locations). The increase of energy consumption on Taipei compared to the other locations leads to lower score for VGG19.

On the contrary,  we observe a different trend when we analyze the models trained on MNIST at Oregon and South Carolina: VGG16 has the highest Score value and, at the same time, the highest accuracy and lowest energy.

The reason why the levels of energy consumption change for each location given identical specifications and experiments is due to how CodeCarbon estimates the net carbon intensity. For each location, the proportion of energy derived from fossil fuels and low-carbon sources are approximated using the international energy mixes derived from the United States' Energy Information Administration's Emissions \& Generation Resource Integrated Database (eGRID). This approximation is done by examining the share of total primary energy produced and consumed for each country in the dataset and determining the proportion of energy derived from different types of energy sources (e.g., coal, petroleum, natural gas and renewables).

By choosing the architecture with highest Score, we obtain either 
(i) an improvement in both accuracy and energy efficiency (e.g., models trained using MNIST dataset), or 
(ii) an improvement in energy efficiency with a detriment (small such in the case on CIFAR10 on South Carolina) on accuracy.

\begin{table}[h]
\caption{Scores of the different experiment configurations. Accuracy: validation accuracy from last epoch of training. Energy: Kilowatt per hour. Score = Accuracy/Energy.}
\resizebox{\columnwidth}{!}{%
\centering
\begin{tabular}{@{}llllll@{}}
\toprule
\textbf{Location} & \textbf{Data} & \textbf{Architecture} & \textbf{Accuracy} & \textbf{Energy} & \textbf{Score} \\ \midrule
Oregon & CIFAR10 & VGG16 & 0.6189 & 0.0583 & 10.63 \\
 &  & VGG19 & 0.6018 & 0.0493 & \textbf{12.64} \\
 &  & ResNet50 & 0.3021 & 0.1057 & 4.11 \\
 & MNIST & VGG16 & 0.9429 & 0.0879 & \textbf{11.02} \\
 &  & VGG19 & 0.9395 & 0.0932 & 10.44 \\ 
  &  & ResNet50 & 0.8858 & 0.1893 & 7.64 \\
 \midrule
S.Carolina & CIFAR10 & VGG16 & 0.6167 & 0.0667 & 9.26 \\
 &  & VGG19 & 0.6157 & 0.0574 & 10.88 \\
 &  & ResNet50 & 0.1 & 0.1224 & 1.17 \\
 & MNIST & VGG16 & 0.9459 & 0.0920 & 10.42 \\
 &  & VGG19 & 0.9384 & 0.1137 & 8.26 \\
 &  & ResNet50 & 0.8883 & 0.2171 & 6.36 \\ \midrule
Taipei & CIFAR10 & VGG16 & 0.6191 & 0.0567 & 10.99 \\
 &  & VGG19 & 0.6147 & 0.0637 & 9.80 \\
 &  & ResNet50 & 0.2169 & 0.1347 & 2.48 \\ \bottomrule
\end{tabular}}
\label{tab:scores}
\end{table}

\subsection{What are the differences between software-based and hardware-based methods of measuring
the energy efficiency of a model? (RQ3)}\label{subsec:rq1}

\begin{table}[h]
\caption{Energy consumption obtained using a wattmeter and a profiler, expressed in kWh. For the profiler, we present the energy consumption and the total reported by the profiler.}
\resizebox{\columnwidth}{!}{%
\centering
\begin{tabular}{@{}llcrrrr@{}}
\toprule
\multirow{2}{*}{\textbf{Data}} & \multirow{2}{*}{\textbf{Archit.}} & \multicolumn{1}{c}{\multirow{2}{*}{\textbf{\begin{tabular}[c]{@{}l@{}}Watt. \\ (kWH)\end{tabular}}}} & \multicolumn{4}{c}{\textbf{CodeCarbon (kWH)}} \\ \cmidrule(l){4-7} 
 &  & \multicolumn{1}{c}{} & \textbf{CPU} & \textbf{GPU} & \textbf{RAM} & \textbf{TOTAL} \\ \midrule
MNIST & VGG16 & 2.25 & 0.04 & 0.77 & 0.41 & 1.21 \\
 & VGG19 & 2.54 & 0.09 & 0.85 & 0.47 & 1.40 \\
 & ResNet50 & 3.03 & 0.05 & 1.08 & 0.59 & 1.72 \\
CIFAR10 & VGG16 & 1.48 & 0.06 & 0.52 & 0.28 & 0.86 \\
 & VGG19 & 1.73 & 0.05 & 0.61 & 0.32 & 0.98 \\
 & ResNet50 & 1.70 & 0.01 & 0.64 & 0.35 & 0.99 \\ \bottomrule
\end{tabular}}
\label{tab:wattcc}
\end{table}

Table \ref{tab:wattcc} shows the median energy consumption obtained using a wattmeter and a profiler. All the values are expressed in kWh. 

We observe that the energy consumption returned by the wattmeter is larger than the total energy consumed reported by the profiler in the same amount of time, going from 42\% to 46\%.
We provide two possible explanations for this difference. 
First of all, a profiler is not analog to a power meter.
The profiler we use, CodeCarbon, is based on RAPL, which uses a software power model which estimates energy usage by using hardware performance counters and I/O models. 
On the contrary, the wattmeter does not estimate the consumption; it actually reports samples of power consumed by the devised connected to it, and from those samples we computed the energy consumed.

Secondly, the total energy computed from the wattmeter also includes the energy consumed by all components from devices connected to the wattmeter (e.g., cache memories, hard disks). On the contrary, the profiles computes the total energy based on estimation from three components: CPU, GPU and RAM consumption.
Nevertheless, we observe a correlation between the energy consumed reported by the wattmeter and by the profiler:
Even the energy values are different, the correlation computed using Spearman gives a rho equals to 0.94, which means a strong correlation.

For example, the architecture ResNet50 on MNIST is, according with both wattmeter and profiler, the configuration that consumes more energy, while VGG16 on CIFAR is the configuration that consumes less.

\section{Discussion}\label{sec:discussion}


Our results show that the selection of different CNN architectures for image classification and dataset size affect the energy consumption as well as the Score (accuracy/energy). This is mainly due to the duration of the training: the larger the number of parameters to train, the longer it will take and consequently the larger the energy consumption. Also, in terms of carbon footprint, the location of the computing infrastructure plays an important role because of the sources of power. These results indicate these two factors as promising to achieve greener DL solutions. Specifically, our RQ2 shows the potential of optimized learning processes requiring less input (data efficient DL) without degrading the quality of the output. With respect to related work, it becomes necessary to explore DL methods dealing with lower volumes such as transfer learning and model compression for more computationally efficient models. Ofori et al. have several works showing that models with pre-trained weights outperform state-out-the-art CNNs \cite{hicss21ofori2021approach, hicss22ofori2022deep}.

Furthermore, profilers are a good estimation of the real energy consumption. The energy profiles are based on software metrics, and provide an estimation of the energy consumed during the training of a model.

Consequently, the energy values obtained with software-based tools are not as precise as those that can be computed using a hardware-based device such as a wattmeter.
However, this study shows that there is a strong correlation between the energy reported by the wattmeter and the energy reported by a profiler. Meaning that, profilers are a cheap (no additional hardware is required), easy (few lines of code are required) and reliable way to compare the energy consumption at the expense of some precision in the calculation.

In this study, we have compared the trade-off between the performance of a model in terms of accuracy and in terms of energy consumption. With this, we have seen that by choosing models that are more energy efficient we are compromising the accuracy of the model, and that little gains of improvement require much more computation. By considering one factor over the other when considering models, one can argue that only when facing critical cases (e.g., medical imaging) these improvements in the model's performance are needed.

The outcomes of this study have provided insight to the process of training a ML model as a 'one-time' operation. However, the energy concerns are raised when we start to include ML in the development and operation chains to Machine Learning Operations (MLOps). As the training and deployment of ML models are automated procedures that are re-trained, updated and maintained in cycle.

\subsection{Limitations}

We faced several threats to validity of our study, for which we took mitigation actions as described below.

{\bf Number of executions}. A single execution in a given configuration may always suffer from some malfunctioning. Therefore, we executed our experiment three times per each configuration and took the mean and median.

{\bf Location}. We could not select the locations beforehand; they were random as Kaggle selected internally the servers used. However, this randomness did not interfere with the execution of the study and the analysis of its results.

{\bf Generalization}. Our results apply for the CV domain, even for two particular datasets used, and cannot be generalized beyond this point without further studies.

{\bf Reliability}. We observed that CodeCarbon yielded occasionally as output negative values of energy consumption.
We conjecture that it can be caused by a bug in the CodeCarbon tool. Nevertheless, to avoid such values impact on our findings, 
we report \emph{median} energy consumption (recall we execute three times each experiment), which means that extreme values (such as those negative) are discarded.
Moreover, we observe that the executions of the ResNet50 architecture trained with CIFAR10 at South Carolina, returned an accuracy of 0.1, and that value did not change along the training process. That could be caused by the malfunction of the Keras platform at that time and location.

\section{Conclusions and future work}\label{sec:conclusions}

In this paper, we have studied three different CNN architectures over two large image classification datasets in order to empirically evaluate the impact of the experimental design in the energy efficiency of the training process. 

Each training session was evaluated with respect to three efficiency metrics: CO$_2$ emissions produced, total energy consumed and number of FLOPs needed. Overall, we gathered statistical evidence of relations between all of the aforementioned variables.

In detail, we have gained statistical evidence that the carbon emissions and the energy consumed by a computational process such as the training of a CNN is related to the experimental design regarding the neural network architecture. We also seen that the impact of the computations can be affected by factors that can be hardly controlled by researchers when engaging in deep learning research, such as the location where the cloud is hosted. 

It is important that the progress of DL research towards better performing models also consider reducing the computational cost of the training when designing the architecture.

Our future work spreads over several dimensions. We plan to replicate the study to other domains, e.g. natural language processing, where other DL architectures may prevail. Also, we aim at further elaborating the results of RQ2 to further extend this notion of score for evaluating models and to address in detail when to stop trading-off the amount of carbon footprint of the model for more accuracy we want to continue studying the variables that are taken into account in the computation of the score. With this, we aim to provide guidance in the creation of future models to obtain better results with less energy consumption.

\section*{Acknowledgments}
This work has been supported by the Spanish project PID2020-117191RB-I00 funded by MCIN/AEI/10.13039/501100011033. 

\section*{Data availability}

The replication package is available on:
\url{https://zenodo.org/badge/latestdoi/503292169}


\bibliographystyle{ieeetr}
\bibliography{hicss}

\end{document}